\pdfoutput=1
\documentclass[journal]{IEEEtranTIE}
\usepackage{graphicx}
\usepackage[numbers]{natbib}
\usepackage{picinpar}
\usepackage{amsmath}
\newcommand\numberthis{\addtocounter{equation}{1}\tag{\theequation}}
\usepackage{amssymb}
\usepackage{tikz}

\ifCLASSOPTIONcompsoc
    \usepackage[caption=false, font=normalsize, labelfont=sf, textfont=sf]{subfig}
\else
\usepackage[caption=false, font=footnotesize]{subfig}
\fi

\usepackage{url}
\usepackage{flushend}
\usepackage[latin1]{inputenc}
\usepackage{colortbl}
\usepackage{soul}
\usepackage{multirow}
\usepackage{pifont}
\usepackage{color}
\usepackage{alltt}
\usepackage[hidelinks]{hyperref}
\usepackage{enumerate}
\usepackage{siunitx}
\usepackage{breakurl}
\usepackage{epstopdf}
\usepackage{pbox}
\usepackage{mathtools}
\usepackage{bm}

\usepackage{algorithm}
\usepackage{algcompatible}
\usepackage{algpseudocode}
\usepackage{ifthen}

\DeclareMathOperator*{\argmax}{arg\,max}
\DeclareMathOperator*{\argmin}{arg\,min}

\newboolean{extended}
\setboolean{extended}{true}
\setboolean{extended}{false}

\newcommand\copyrighttext{%
  \footnotesize This work has been submitted to the IEEE for possible publication. Copyright may be transferred without notice, after which this version may no longer be accessible.}
\newcommand\copyrightnotice{%
\begin{tikzpicture}[remember picture,overlay]
\node[anchor=south,yshift=10pt] at (current page.south) {\fbox{\parbox{\dimexpr\textwidth-\fboxsep-\fboxrule\relax}{\copyrighttext}}};
\end{tikzpicture}%
}

\begin{document}

\title{Remaining Useful Life Estimation Under Uncertainty with Causal GraphNets}

\author{
	\vskip 1em
	
	Charilaos Mylonas, 
	Eleni Chatzi

	\thanks{
	
		Manuscript received Month xx, 2xxx; revised Month xx, xxxx; accepted Month x, xxxx.
		This work was supported by ERC Starting Grant (ERC-2015-StG \#679843) on the topic of  "Smart Monitoring, Inspection and Life-Cycle Assessment of Wind Turbines" at ETH Z\"urich.
	  Charilaos Mylonas and Eleni Chatzi are with the Department of Civil and Environmental Engineering, ETH Z\"urich, Zurich 8093, Switzerland (e-mail: 
    \texttt{charilaos.mylonas@gmail.com}, \texttt{chatzi@ibk.baug.ethz.ch}). 
		
	}
}

\maketitle
	
\begin{abstract}

In this work, a novel approach for the construction and training of time series models
is presented that deals with the problem of learning on large time series with non-equispaced 
observations, which at the same time may possess features of interest that span multiple scales. 
The proposed method is appropriate for constructing predictive models for non-stationary stochastic time series.
The efficacy of the method is demonstrated on a simulated stochastic degradation dataset and on a real-world
accelerated life testing dataset for ball-bearings.
The proposed method, which is based on GraphNets,
implicitly learns a model that describes the evolution of the system 
at the level of a state-vector rather than of a raw observation. 
The proposed approach is compared to a recurrent network with a temporal convolutional feature extractor head
(RNN-tCNN) which forms a known viable alternative for the problem context considered.
Finally, by taking advantage of recent advances in the computation of reparametrization 
gradients for learning probability distributions, a simple yet effective technique 
for representing prediction uncertainty as a Gamma distribution over remaining 
useful life predictions is employed.
\end{abstract}


\markboth{PREPRINT SUBMITTED TO IEEE JOURNAL}
{}

\definecolor{limegreen}{rgb}{0.2, 0.8, 0.2}
\definecolor{forestgreen}{rgb}{0.13, 0.55, 0.13}
\definecolor{greenhtml}{rgb}{0.0, 0.5, 0.0}

\section{Introduction}
\copyrightnotice

\IEEEPARstart{P}{redictive} tasks relying on time series data form a focal area overarching diverse technological and scientific fields.
  Settings where observations are available in non-equispaced and sparse intervals require approximations on the evolution of 
  the time series. 
	Physics-based models able to simulate the evolution of a system could offer such predictive capabilities, 
  but are typically either unavailable, of lower precision, or associated with prohibitively expensive 
  numerical computations and/or modeling effort. On the other hand, there exist settings,
  where readily available measurement data correlate in a non-trivial manner with quantities of interest. 
  Moreover, when the evolution of the system at hand is non-deterministic, 
  even if a perfect knowledge of the instantaneous system state is somehow achieved, a
  deterministic estimate of the long-term evolution of the system is not possible.
  Therefore, it is of utmost importance to represent the uncertainty in the predictions involving stochastically 
  evolving systems. This work focuses on the problem of Remaining Useful Life (RUL) prediction, which 
  encompases these characteristics.

  In many real-world applications, as in the case-study examined herein, a model of degradation and final failure 
  is not available or not reliable enough\footnote{At this point a clear distinction of the model of degradation of a component and a model of the time series of the component should be made. We consider settings where we have neither but have raw measurements of the latter.}.
  Therefore, such a model has to be learned directly from field or experimental observations. Although the physics of the considered problem are relatively
  well understood, the uncertainty in various parameters involved in analyzing such systems, such as geometric deviations, effect of 
  environmental conditions on lubricant properties, material and manufacturing imperfections and the effect of not fully observable loading conditions do not allow for a 
  treatment of the problem where all physical processes are accounted for.
  In the same context, it is expected that features related to the damage of the components evolve \textit{stochastically} 
  and the damage \textit{state} has an \textit{indirect} effect on the observed raw time series. 

  The method proposed herein is inspired by the recent advances in \textit{GraphNets} (GNs) and the flexibility 
  these allow for in terms of defining \textit{inductive biases}. The GraphNets framework, as introduced in \cite{battaglia2018relational}, is a generalization on possible computations 
  on attributed graphs, which covers Graph Neural Network (GNN) techniques, such as Message-Passing Neural Networks (MPNNs) 
  \cite{gilmer2017neural} and Non-local Neural Networks (NLNNs) \cite{wang2018non}.
  An inductive bias (or \textit{learning bias}) is any belief or assumption that, when incorporated 
  in the training procedure, can facilitate a machine learning algorithm to learn with fewer data or better generalize in unseen settings.
  In practice, for the problem of RUL estimation, due to interruptions in transmission or storage limitations, monitoring time series 
  contain gaps \cite{sikorska2011prognostic}. The non-regular sampling of the time series data is routinely treated as a missing data problem;
  a task most commonly referred to as \textit{time series imputation} \cite{razavi2019integrated}.
  This requires to impose an explicit evolution model that reproduces \textit{the raw time series itself} in regular intervals, so that algorithms designed 
  to work with data observed in regular intervals can be used. This approach biases the subsequent treatment of the data with predictive algorithms.
  The present work, in addition to providing a solution to long time series, yields a radically different approach 
  to the problem of non-regular observations for building predictive models.
  Instead of completing the missing data and subsequently employing a time series technique that operates on equispaced data, a model that accumulates the information of 
  the available non-regularly spaced data is learned directly. Instead of an explicit model that reproduces the time series, the temporal ordering
  of the observations is incorporated in the learning algorithm as an \textit{inductive bias}.

  Incorporation of inductive biases is useful in constructing machine 
  learning models that perform well when trained on relatively small datasets and for building smaller and more computationally efficient models.
  Recurrent neural networks (RNNs) impose a chain-structure of dependence, which constrains RNN algorithms to sequential computations and typically require $N$ sequential steps of computation 
  to propagate information from observations that lie $N$ steps before the current time step. When considering very large $N$, this becomes a significant computational disadvantage 
  both in training and evaluation of RNNs for long time series. Other recent original approaches to sequence modeling, such as NeuralODEs \cite{chen2018neural} and Legendre
  Memory Units \cite{voelker2019legendre} offer a solution to the issue of non-equispaced data, but do not facilitate the easier propagation of information from arbitrary past steps since they 
  retain the chain structure of RNNs.
  In contrast, the architecture proposed in this work does not assume a chain graph for processing the past time-steps but a more general causal graph. 
  Thus, the proposed architecture can learn in a more parallelized manner with a constant (and adjustable) 
  number of sequential computational steps, as will be detailed later in the text.

  
  \ifthenelse{\boolean{extended}}{
  A model of the evolution of the system is learned in an implicit manner, as the evolution 
  of a \textit{latent space} that affects in a potentially non-trivial manner the raw observations.
  Instances of the aforementioned latent space, provide evidence on the instantaneous state of the system.
  The predictive algorithm proposed does not use this estimate directly, but accumulates 
  evidence from available past observations in order to yield more accurate results that take into 
  account the evolution of the system due to operational conditions, which, for the purposes of this study, 
  are assumed known as it is often the case in practice for continuously monitored components.
  The effect of the operational conditions to the evolution of the state-space is not assumed known, 
  but is also learned directly from data through parametrizing the so-called edge-function of the GraphNet.
  The exact manner this is achieved is detailed in \autoref{sec:gnncomputation}.}
{}


  \ifthenelse{\boolean{extended}}
  {During the last decade it has been well established by numerous successful demonstrations, particularly in computer vision \cite{simonyan2014very}, that feature extraction can be performed in tandem 
  with learning the predictive model. This allows for more principled and less arduous single-step training as compared to the feature extraction \& discriminative model learning. 
  In machine learning for time series models, it is mainly generative models \cite{wavenet} that have shown promise in training end-to-end with gradient descent.
  Works on end-to-end trainable discriminative models have been limited. 
  }{}
  Classical machine learning techniques for general sequence datasets consist of separate feature extraction \& selection and predictive model construction and selection pipelines. 
  The most widely used feature extraction techniques, naturally fitting to time series models, are (1) Discrete Fourier Transforms (DFT), due to the intuitive decomposition of 
  the signal to coefficients (2), Wavelet transforms, owing to the multi-scale time-frequency characteristics of some signals, and (3) Dynamic Time-Warping (DTW),
  when the main source of variation among signals is due to some temporal distortion (i.e. non-stationarity), such as different heart-rates in EEG
  classification \cite{raghavendra2011cardiac} or different rotational speeds in machinery \cite{zhen2013fault}. In several applications of machine learning for predictive 
  time series models simple moments of the signals are used, such as kurtosis and standard deviation of time series segments \cite{rouet2017machine}.
  In several application fields, special expert-guided feature extraction techniques have been proposed to facilitate downstream tasks. One successful representative 
  example of this class of models in time series analysis are mel-cepstral features \cite{kitamura1990speaker} in human speech and music processing. 
  For most other applications, the classical machine learning workflow is followed, where a large set of features are pre-computed and, in a second stage, features are selected 
  by inspecting the generalization performance of the model (for instance with cross-validation). When physical intuition is not easy to draw from for 
  the problem at hand, features are extracted by unsupervised learning techniques \cite{yang2018automatic}, 
  such as autoencoders, or special negative-sampling based losses, such as time-contrastive learning \cite{hyvarinen2016unsupervised}.
  Combinations of unsupervised learning techniques (such as autoencoders and deep Boltzmann machines)
  and hand-crafted pre-processing with DCT are also used \cite{deng2010binary}.

  A number of works apply deep learning for the RUL prediction problem from time series data. In \cite{yang2019remaining} two CNN-based 
  predictors are trained. One classifier predicts the point in time where the sudden increase in the amplitude of accelerations
  occurs which is close to failure and subsequently a second classifier predicts the time-to-failure after that point. The same approach is followed in \cite{shi2020remaining},
  where Random Forests and XGBoost are used as predictive models.
  In \cite{wang2020recurrent}, a recurrent convolutional network is adopted \cite{liang2015recurrent} and Monte-Carlo Dropout \cite{gal2016dropout} is used
  as a simple and effective way of representing the uncertainty in the predictions. In \cite{wang2020multi}, instead of recurrent connections, as applied
  in \cite{liang2015recurrent}, \textit{attention} layers are used to enhance the performance of CNNs.
  All aforementioned approaches are not appropriate for arbitrarily spaced data, as there is no explicit representation of the time between the observations.
  In the present work a uniform treatment of the different stages of degradation, is proposed without attempting to classify different stages of degradation since they are not 
  clearly defined and this approach could bias unfavorably the results.

  \ifthenelse{\boolean{extended}}{
  In summary, the novelties of this work are as follows:
    \begin{itemize}
      \item the inductive biases related to the causal structure of the observed time series and potential forcing of the evolution of the time series are represented by a learnable state-evolution model
      \item the (non-stationary) evolution of the time series is treated as evolution of an (implicit) latent state-space, as opposed to the evolution of the raw observations of the time series,
      \item the function relating observations of segments of the time series to the (implicit) state-space and the function describing the evolution of the time series is learned in an end-to-end differentiable manner, allowing for mini-batched training (i.e. with stochastic gradient descent) and therefore favorable scaling properties when dealing with large datasets.
    \end{itemize}
  }{}

\section{Degradation Time Series Datasets}
\subsection{A Simulated Degradation Process Dataset}
\begin{figure}[h!]
	\center
	\includegraphics[width=0.45\textwidth]{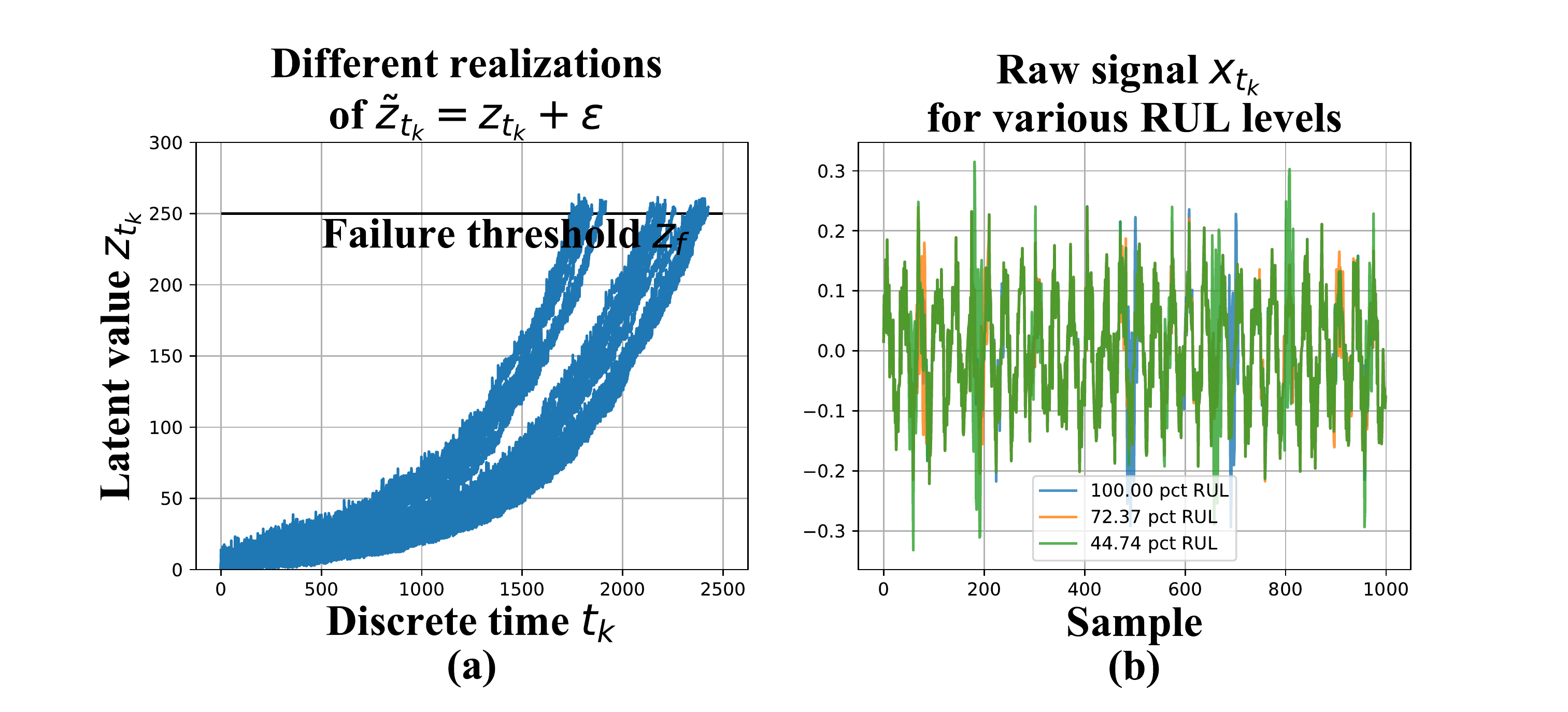}
  \caption{(a) Simulated latent variable $z_{t_k}$ and (b)
  raw high-frequency time series realizations $x_{t_k}$.}
  \label{fig:latentfict}
\end{figure}

In order to verify the efficacy of the method for remaining useful life prediction tasks over long time series, a non-stationary degradation process was simulated.
The underlying process governing the degradation is a non-stationary Markov process with Gamma distributed increments \cite{bogdanoff}. The parameters of the
Gamma distributions producing the increments are assumed to depend on the previous steps, since damage propagation does depend on previous damage 
states. In physical terms, this simulates the path dependence of irreversible processes. The random process presented herein does not have a direct physical 
analog and is only designed to demonstrate the properties of the proposed algorithm. The process generating the latent space as follows
\begin{align*}
  \delta \eta_{t_i}^{(\alpha,\beta)} & \sim Gamma(\alpha(t_i,c), \beta),\quad \alpha(t,c) = 0.02 + t^c\\ 
  z_{t_k}^{(\alpha,\beta)} & = \sum_{i=0}^{t_k} \delta \eta_{t_i}^{(\alpha, \beta)},\quad z_{t_k}^{(\alpha,\beta)}  < z_f \numberthis 
\end{align*}
where $ \{t_0, t_1, \cdots t_N \} $ are consecutive, discrete time steps, $\eta_t^{(\alpha,\beta)}$ is a random variable with a non-linear dependence on time and $c$ a random variable different for each experiment.
The parameters $\alpha > 0 $ and $\beta > 0 $ are termed the \textit{concentration or shape} and \textit{rate} parameters of the Gamma distribution.
The probability density function of a Gamma distribution is defined as $f(x;\alpha,\beta) = \frac{\beta}{\Gamma(\alpha)}x^{\alpha-1} e^{-x \beta}$ where $\Gamma(\cdot)$ 
is the Gamma function.
Failure occurs when the latent accumulating damage variable $z_{t_k}$ reaches a threshold value, which is the same for all experiments, denoted as $z_f$.
It is assumed that the different experiments have slightly different evolutions for their damage, which may arise from variations in manufacturing.
This is simulated by sampling $c$ from a Gaussian distribution.
The non-linear dependence is realized through the shape parameter $\alpha(t,c)$ of the Gamma distribution controlling the size of the increments. 
It should be noted that the non-linear dependence on time is used to simulate the non-stationarity of the process, due to dependence of ``$\alpha$''
on the accumulated $z_{t_k}$. 
The high-frequency instantaneous measurement of the signal is denoted as $x_{t_k}$.
The observations of the process consist of 1000 samples that contain randomly placed spikes with amplitude that non-linearly depends on $z_{t_k}$, a process denoted with $G(\cdot)$ for conciseness.
\begin{equation*}
\begin{aligned}[c]
  \tilde{z}_{t_k} & =  z_{t_k} + \epsilon \\ 
  x_{t_k} &= G(\tilde{z}_{t_k}) + \zeta \\ 
\end{aligned}
\qquad 
\begin{aligned}[c]
  \epsilon & \sim \mathcal{N}(0,{\sigma_z}^2) \\ 
  \zeta& \sim \mathcal{N}(0,{\sigma_{x}}^2) \\ 
\end{aligned}
\end{equation*}
%
Gaussian noise is added both to the raw signal observation $x_{t_k} $ and directly to the latent variable $z_{t_k}$. Noise $\zeta$ is the
observation noise. Noise $\epsilon$, is added to the instantaneous latent damage state $z_{t_k}$ in order to model the fact that $z_{t_k}$ 
may not be accurately determinable from $x_t$ even in the absence of $\zeta$.
Each process underlying the observations of each experiment, evolves in the long-term in a similar yet sufficiently varied manner as shown in \autoref{fig:latentfict}.
A set of $x_t$ signals (raw observations) are shown in \autoref{fig:latentfict}(b). 
Although this process does not correspond directly to some actual physical problem, it is argued that it possesses all the necessary 
characteristics of a prototypical RUL problem and a useful test-case.
\subsection{An experimental dataset on accelerated fatigue of ball bearings}
 
\ifthenelse{\boolean{extended}}{
  We here adopt the experimental platform used in \cite{nectoux2012pronostia} which is schematically depicted in \autoref{fig:pronostia}.
  The dataset was introduced as part of the 
  2012 Prognostics and Health Management conference challenge. The platform supports loading perpendicular to the axis of rotation 
  and rotation at different speeds. Measurements of temperature and acceleration are being acquired during the accelerated testing procedure.
}
{
} 

\ifthenelse{\boolean{extended}}{
\begin{figure}[h!]
	\centering
	\includegraphics[width=0.45\textwidth]{pronostia_rig_drawing.pdf}
	\caption{A schematic of the accelerated fatigue testing rig used. The lateral force $F_i$ and the rotation speed $\dot \phi_i$ have values shown in \autoref{tab:conditions} ($i \in \{A, B, C\})$. More details on the experimental setup can be found in \cite{nectoux2012pronostia}}
	\label{fig:pronostia}
\end{figure}
}{
}
\ifthenelse{\boolean{extended}}{}{
  The dataset in \cite{nectoux2012pronostia} consists of run-to-failure experiments of a total of 17 bearings, loaded in 3 different conditions. Only 2-axis acceleration measurements are used in the present work. Temperature measurements are also available.
}Importantly,
no artificial damage is introduced to the components for accelerating failure, thus rendering the accelerated testing scenario a better representation of
real-world settings, where the failure mode is not known a-priori.
The experimental conditions are summarized in \autoref{tab:conditions}. 

In order to test generalization on un-seen experiments a test-set containing whole experiments is used. 
A different train/test split is adopted from \cite{nectoux2012pronostia}, which is detailed in \autoref{tab:traintest}.
\begin{table}[h!]
    \centering
    \caption{Available experiments and loading conditions.}
    \label{tab:conditions}
    \begin{tabular}{l l l l}
      \hline
      \hline
    	Conditions $i$ & $\dot \phi_i$ [rpm]   & $F_i$ [kN] & Number of experiments\\
      \hline
      A & 1800 & 4.0 & 7\\
      B & 1650 & 4.2 & 7 \\
      C & 1500 & 5.0 & 3 \\
      \hline
      \hline
    \end{tabular}
    \caption{FEMTO bearings dataset, training/test split} 
	\label{tab:traintest}
    \begin{tabular}{l l l l l}
      \hline
      \hline
      Set        & Experiment    & Conditions & Failure time [s] & Num. Obs.\\
      \hline
     Training   & 1\_2           & A          & 8700             & 871 \\
                & 1\_3           & A          & 23740            & 2375 \\
                & 1\_4           & A          & 14270            & 1428 \\
                & 1\_5           & A          & 24620            & 2463 \\
                & 2\_1           & B          & 9100             & 911 \\
                & 2\_5           & B          & 23100            & 2311 \\
                & 2\_6           & B          & 7000             & 701 \\
                & 3\_3           & C          & 4330             & 434 \\
  \hline \\
     Testing  & 1\_1             & A          & 28072            & 2803 \\
              & 1\_6             & A          & 24470            & 2448 \\
              & 1\_7             & A          & 22580            & 2259 \\
              & 2\_2             & B          & 7960             & 797 \\
              & 2\_3             & B          & 19540            & 1955 \\
              & 2\_4             & B          & 7500             & 751 \\
              & 2\_7             & B          & 2290             & 230 \\
              & 3\_1             & C          & 5140             & 515 \\
              & 3\_2             & C          & 16360            & 1637 \\
      \hline
      \hline
  \end{tabular}

\end{table}

Fatigue damage on roller bearings, manifests as frictional wear of the
bearings and/or the surrounding ring. 
\ifthenelse{\boolean{extended}}
{Empirically, it is expected that 
the time-to failure $y$ will follow $y \sim \frac{1}{\dot \phi_i} \cdot \frac{1}{F_{i}^3}$ where $F_i$ is 
the lateral applied load and $\dot \phi_i$ is the  rotational speed of the bearings. 
Failure is expected to initiate due to load concentrations attributed 
to contamination or manufacturing imperfections. The lateral load applied, 
magnifies the stress concentrations due to these imperfections and consequently causes
failure to occur faster. As observed in \autoref{tab:traintest} there is 
only an approximate trend with respect to the time-to-failure and higher 
lateral loads. Moreover, as it is the norm with fatigue related failure, 
there is a large scatter in the remaining time to failure of the 
experiments with identical lab-controlled loading conditions.}
{Empirically, higher lateral loads $F_i$ and rotational speeds $\dot \phi_i$ are associated with faster wear for bearings.}
The $\ddot u_x$ and $\ddot u_y$ accelerometer data are available in 0.1 second segments,
sampled at 25.56kHz (2556 samples per segment). 
Temporal convolution networks are used, to automatically detect and use features that potentially are useful to tracking degradation in these experiments.

\section{Model Architectures}
\subsection{GraphNets for arbitrary inductive biases}
GraphNets (GNs) are a class of machine learning algorithms operating with 
(typically pre-defined) attributed graph data, which generalize 
several graph neural network architectures. 
An attributed graph, in essence, is a set of
nodes (vertices) $V : \{\mathbf{v}_{1}, \cdots \mathbf{v}_{k} \}$ and 
edges $E : \{(\mathbf{e}_{1}, r_{1},s_{1})\cdots (\mathbf{e}_{k}, r_k, s_k) \} $ where $\mathbf{e}_{k} \in \mathbb{R}^{N^e}$ and $\mathbf{v}_{i} \in \mathbb{R}^{N^v}$.
Each edge is a triplet $(\mathbf{e}_j, r_j, s_j)$ (or equivallently $(\mathbf{e}, \mathbf{v}_{r_j}, \mathbf{v}_{s_j})$) and it contains a reference to a receiver
node $\mathbf{v}_{r_j}$, to a sender node $\mathbf{v}_{s_j}$ as well as a (vector) attribute $\mathbf{e}_j$.
Self-edges, i.e. when $r_i := s_i$ are allowed. 
In \cite{battaglia2018relational} a more general class of GraphNets is presented where \textit{global} variables which affect all nodes and edges are allowed.
A GN with no global variables consists of a node-function $\phi^v$, an edge function $\phi^e$, and an edge aggregation function $\rho^{e \rightarrow v}$. 
The function $\rho^{e \rightarrow v}$ should be \textbf{(1)} invariant to the permutation of its inputs and \textbf{(2)} able to
accept a variable number of inputs. In the following this will be referred to as the \textit{edge aggregation function}. 
Simple valid aggregation functions are $Min(\cdot)$, $Max(\cdot)$, $Sum(\cdot)$ and $Mean(\cdot)$. 
Inventing more general aggregation functions (for instance by combining them) and investigating 
how they affect the approximation properties of GNs is an active current research subject \cite{corso2020principal}.

Ignoring global graph attributes, the GraphNet computation procedure is as detailed in algorithm \autoref{alg:gnnblock}.
First, the new edge states are evaluated using the sender and receiver vertex attributes ($\mathbf{v}_{s_i}$ and $\mathbf{v}_{r_i}$ correspondingly) and the previous edge state $\mathbf{e_i}$ as arguments to the edge function $\phi^e$.
The arguments of the edge function may contain any combination of the source and target node attributes and the edge attribute.
Afterwards, the nodes of the graph are iterated and the incoming edges for each node are used to compute an aggregated incoming edge \textit{message} $\mathbf{\bar{e}}'_i $
The aggregated edge message together with the node attributes are used to compute an \textit{updated} node state.
Typically, small Multi-Layer Perceptrons (MLPs) are used for the edge and node GraphNet functions $\phi^e$ and $\phi^v$.
It is possible to compose GN blocks by using the output of a GN as the input to another GN block.
Since a single GN block allows only first order neighbors to exchange messages, GN blocks are composed as
\begin{align*}
  GN_K(GN_{K-1}(\cdots(GN_0(G)\cdots)))  = \\
  GN_K \circ GN_{K-1} \circ \cdots \circ GN_0(G)
\end{align*}
where ``$\circ$'' denotes composition.
The first GN block may cast the input graph data to a lower dimension so as to allow for more efficient computation.
The first GN block may have edge functions that depend only on edge states $\phi^{e_0}(\mathbf{e})$ and correspondingly node functions 
that depend only on node states $\phi^{u_0}(\mathbf{v})$. This is refered to a \textit{Graph Independent} 
GN block and it is used as the type of layer for the first and the last GN block. 
The inner GN steps (i.e. $GN_1$ to $GN_{K-1}$) are \textit{full} GN blocks, 
where message passing takes place. This general computational pattern is refered 
to as \textit{encode-process-decode} \cite{battaglia2018relational}. The inner GN 
blocks may have shared weights, yielding smaller memory footprint for the whole model or 
different weights, ammounting to different GN functions that need to be trained for each level. Sharing 
weights and repeatedly applying the same GN block helps propagate and combine information 
from more connected nodes in the graph.
\begin{algorithm}[th!]
\begin{algorithmic}
\Function{GraphNetwork}{$E$, $V$}
    \For {$k\in \{1\ldots{}N^e\}$}
        \State $\mathbf{e}_k^\prime\gets \phi^e\left(\mathbf{e}_k, \mathbf{v}_{r_k}, \mathbf{v}_{s_k}\right)$
        \Comment{1. \small{Compute updated edges} }
    \EndFor
    \For {$i\in \{1\ldots{}N^n\}$}
        \State \textbf{let} $E'_i = \left\{\left(\mathbf{e}'_k, r_k, s_k \right)\right\}_{r_k=i,\; k=1:N^e}$
        \State $\mathbf{\bar{e}}'_i \gets \rho^{e \rightarrow v}\left(E'_i\right)$
        \Comment{2. \small{Aggregate edges per node}}
        \State $\mathbf{v}'_i \gets \phi^v\left(\mathbf{\bar{e}}'_i, \mathbf{v}_i, \right)$
        \Comment{3. \small{Compute updated nodes }}
    \EndFor
    \State \textbf{let} $V' = \left\{\mathbf{v}'\right\}_{i=1:N^v}$
    \State \textbf{let} $E' = \left\{\left(\mathbf{e}'_k, r_k, s_k \right)\right\}_{k=1:N^e}$
    \State \Return $(E', V')$
\EndFunction
\end{algorithmic}
  \caption{GN block without global variables \cite{battaglia2018relational}.}
\label{alg:gnnblock}
\end{algorithm}
%
\ifthenelse{\boolean{extended}}{
There is strong empirical evidence, that good inductive biases allow for training machine learning algorithms that generalize well with fewer data. 
Two of the most successful architectures, namely Recurrent Neural Networks (RNNs) such as Long-Short-Term Memory Networks \cite{lstm} and Convolutional Neural Networks (CNNs)\cite{cnn} 
implement such inductive biases as architectural choices. More speciffically, RNNs implement the causal structure of time series through chain-structured computation 
and CNNs implement the equivariance to translation which is inherited by the convolution operation. }

In the present work, as is the case with RNNs \cite{lstm} and causal CNNs \cite{wavenet}, the causal structure 
of time series is also exploited, which is a good inductive bias for the problem at hand, although \textit{without} requiring that the data 
is processed as a chain-graph or that the data are regularly sampled. Instead, an arbitrary causal graph for the underlying state is built, 
together with functions to infer the quantity of interest which is the remaining useful life of a component 
given a set of non-consecutive short-term observations.


\subsection{Incorporation of Causal Inductive Biases using GraphNets}
\label{sec:gnncomputation}
\begin{figure}[h!]
	\center
	\includegraphics[width=0.35\textwidth]{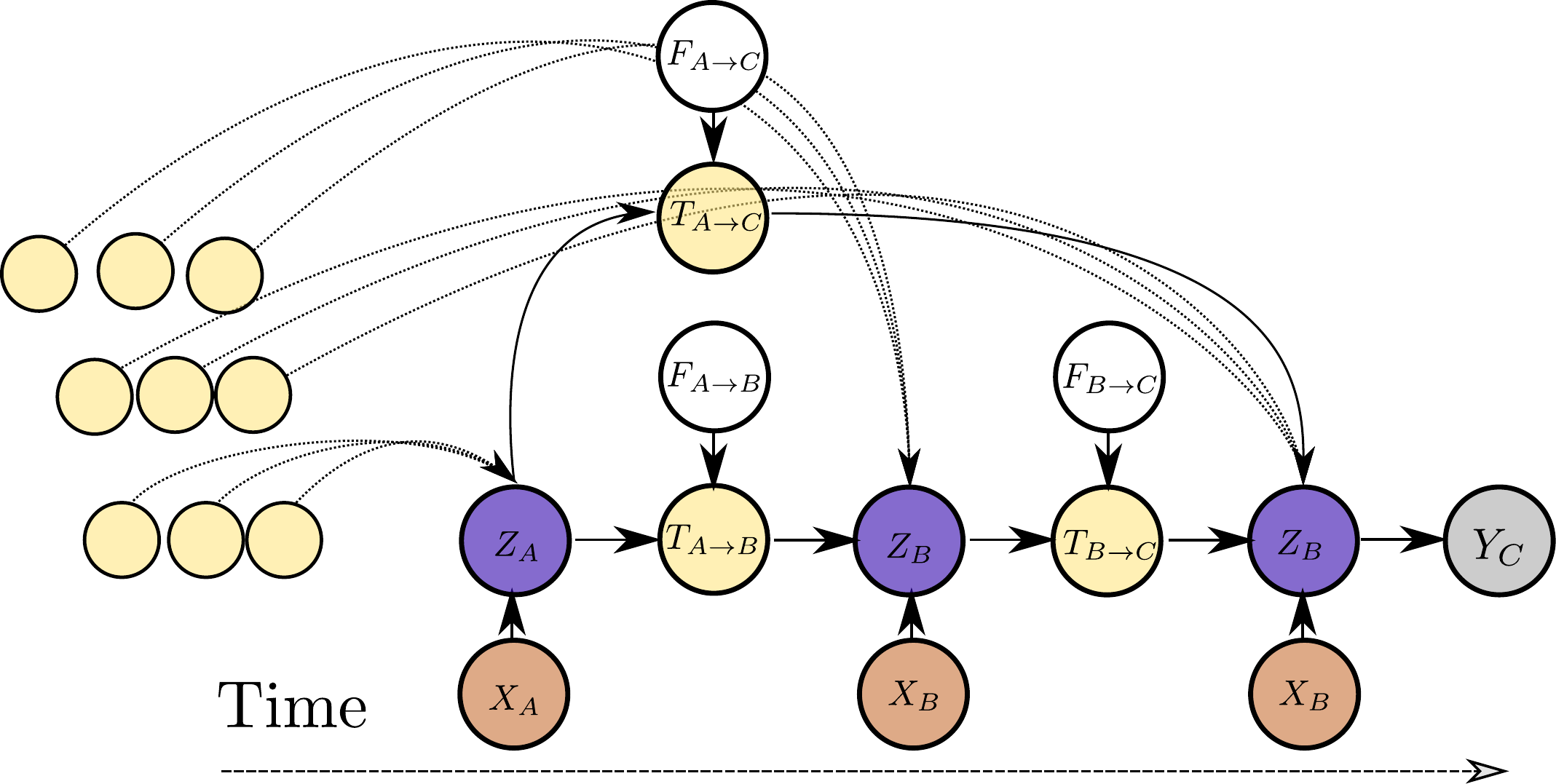}
  \caption{Dependency graph for the variables associated with the proposed model. $X_A$ represents the raw observed variable at time $t_A$. Variable $Z_A$ represents the (unobserved) state that can be translated to the quantity of interest $Y_C$ or a probabilistic estimate.} 
  \label{fig:graphmod}
\end{figure}

\begin{figure}[h!]
  \center
  \includegraphics[width=0.35\textwidth]{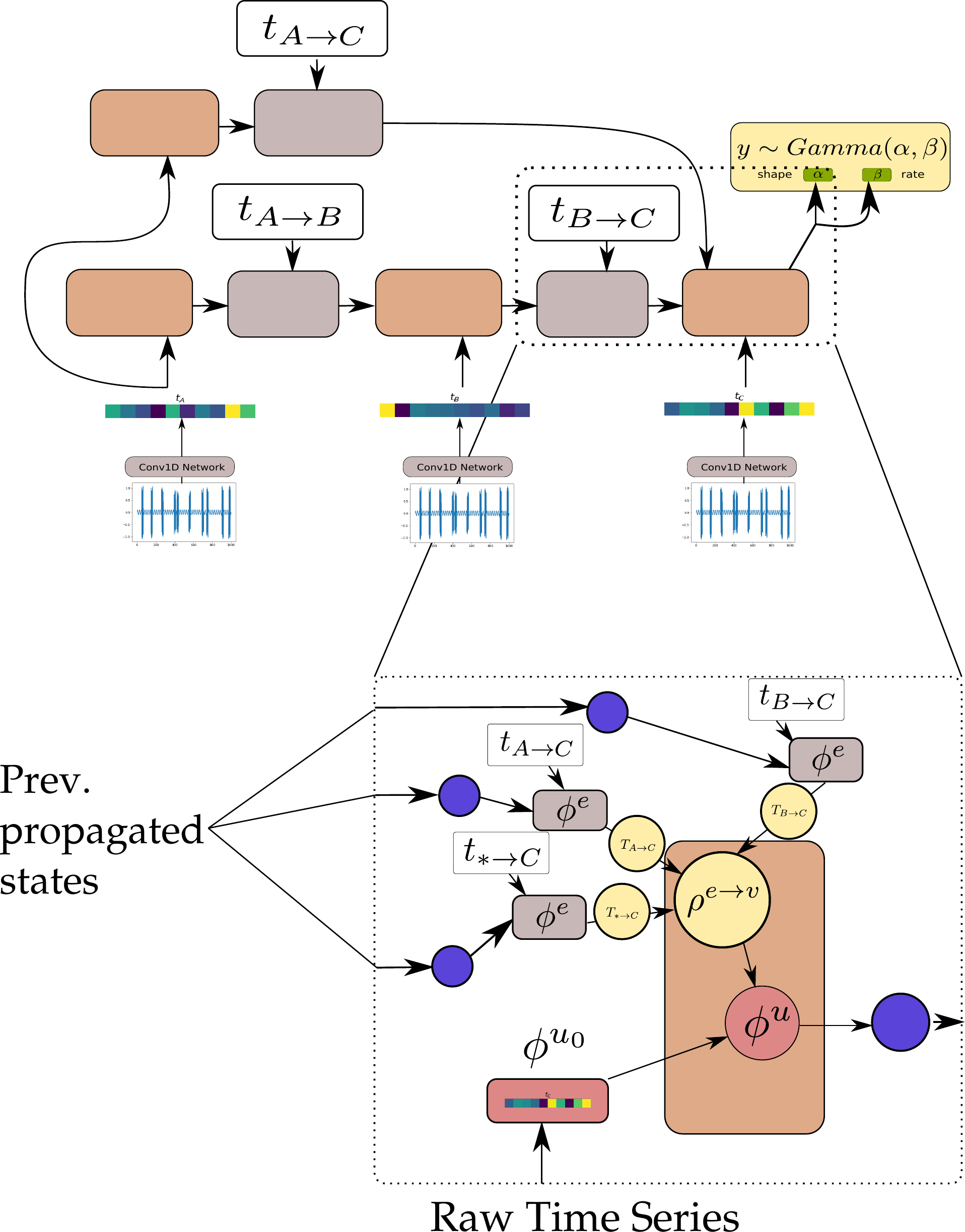}
  \caption{Detail of the \textbf{GNN-tCNN}. States are propagated with $\phi^e$ and accumulated with $\rho^{e \rightarrow u}$.}
  \label{fig:comparch}
\end{figure}

The variable dependencies of the proposed model are schematically depicted in \autoref{fig:graphmod}
for 3 observations. The computational architecture is depicted in more detail in \autoref{fig:comparch}.
The variable $Z_K$ represents the current estimate for the latent state of the system\footnote{It can be considered that the variable contains values that represent, for instance, sufficient statistics or a re-parametrization of a probability distribution, allowing its interpretation as a representation of a probability distribution.}.
The variable $T_{K\rightarrow L}$, which represents
the propagated latent state from past observations, depends on 
the latent state $Z_K$, an exogenous input $F_{K \rightarrow L}$ that 
controls the propagation of state $Z_K$ to $Z_L$ and potentially other propagated latent state estimates from instants before $t_L$.
The exogenous input $F_{K \rightarrow L} $ to the state propagation function can be as simple as the elapsed time between two time 
instants $t_{K\rightarrow L} = t_L- t_K$ or incorporate more prior inductive biases, such as the values 
representing different operating conditions during the interval between observations.
An arbitrary number of past states can be propagated from past observations and \textit{aggregated}
in order to yield better estimates for a latent state $Z_L$. In adition to propagated latent states, instantaneous observations of raw data $X_K$ inform 
the latent state $Z_K$.
For instance, in \autoref{fig:graphmod}
$Z_C$ depends on $T_{B \rightarrow C}$ but at the same time on $T_{A \rightarrow C}$ and potentially 
more propagated states from past observations (other yellow nodes in the graph) and at the same time to an instantaneous observation $X_C$. 
\ifthenelse{\boolean{extended}}{
Note that the final estimate of the quantity of interest $Y_C$, the raw observations $X_K$ and $F_{K \rightarrow L} $ 
have \textit{a direct physical meaning}, whereas $Z_K,\, T_{K \rightarrow L}$ do not necessarily have a direct interpretation.
In what follows, the machine learning architecture that incorporates the aformationed dependencies is described in detail.}
{
  Each inferred latent state $Z_i$ can be transformed to a distribution for the quantity of interest $Y_i$.
}
The value of the propagated state variable from state $s$ to state $d$,  $T_{s \rightarrow t}$, depends jointly on the edge attributes and on the 
latent state of the source node. 
In a conventional RNN model, $ F_{K \rightarrow L}$ corresponds to an exogenous input for the RNN cell.
In contrast to an RNN model, in this work the dependence of the estimate of each state depends on \textit{multiple} states
by introducing a propagated state that is modulated by the exogenous input. In that manner an arbitrary and \textit{variable} number of past states can be used
directly for refining the estimate of the current latent state, instead of the estimate summarized in the latent cell state of the last RNN cell state.
In the proposed model, the parameters of the functions relating the variables of the model are learned directly from data while only defining 
the inductive biases following naturally from the temporal ordering of the observations.
This approach allows for uniform treatment of all observations from the past and allows for the consideration of
an arbitrary number of such observations to yield an estimate of current latent state.

The connections from all observable past states and the ultimate one, where prediction (\textit{read-out}) is performed,
are implemented as a node-to-edge transformation and subsequent aggregations.
Aggregation corresponds to the edge-aggregation function $\rho^{e \rightarrow u}(\cdot)$ of the GraphNet.
In this manner, it is possible to propagate 
information from all distant past states on a single computation step. The computation of all available past states 
would be innefficient. To remedy that, it is possible to randomly sample the past states used 
in order to perform inference for the current step. Similarly, during training it is possible to 
yield unbiased estimates of gradients for the propagation and feature extraction model by randomly sampling 
the past states. It was found that for the presented use-cases this was an effective strategy for training.
\ifthenelse{\boolean{extended}}{
It should be noted that how random sampling from past states is performed for the proposed method 
is a \textit{problem specific} component of the method. The temporal window of the past states considered for sampling should cover 
the longest periods of the time series for the phenomenon of interest. Also a minimum spacing between 
sampled observations is assumed, so that the phenomenon of interest evolves adequately for it to be 
detected. Lastly, it is important that the sampling procedure is the same for training and evaluation so 
the distribution of the input values of the exogenous forcing $F_{i \rightarrow j}$ is similar for training and evaluation.}
{}
In GN terms, the ``\textit{encode}'' GraphNet block ($GN_{enc} : \{ \phi^{u_0}, \phi^{e_0} \}$) is a graph-independent 
block consisting of the node function $\phi^{u_0}$ and edge function $\phi^{e_0}$. The node function is a temporal convolutional 
neural network (temporal CNNs), with architecture detailed in \autoref{tab:tcnn}.

\begin{table}[h!]
    \centering
    \caption{Details on temporal CNN which acts as the node-function $\phi^{u_0}$ of the graph independent $GN_{enc}$  GraphNet. $n^k,n^s,n^f$ corresponds to kernel size, stride and number of filters. For dense layers $n^f$ corresponds to the layer width.}
    \label{tab:tcnn}
    \begin{tabular}{l l l }
      \hline
      \hline
      Layer type  &  $(n^k,n^s, n^f) $   & Activation\\
      \hline
      Conv1D          & $(1 \times 1,1,50)$     & -\\
      Conv1D          & $(1 \times 3,2,18)$     & -\\
      Conv1D          & $(1 \times 3,2,18)$     & Dropout $20\%$ $ReLU$ \\
      Conv1D          & $(1 \times 3,2,50)$     & \\
      Average Pool & $(1 \times 2,2,1)$         & - \\
      \hline

      Conv1D          & $(1 \times 1,1,50)$      & -\\
      Conv1D          & $(1 \times 3,2,18)$      & -\\
      Conv1D          & $(1 \times 3,2,18)$      & Dropout $20\%$ $ReLU$ \\
      Conv1D          & $(1 \times 3,2,50)$      &-\\
      Avg. Pool & $(1 \times 2,2,1)$             & - \\
      \hline

      Conv1D          & $(1 \times 1,1,50)$      & -\\
      Conv1D          & $(1 \times 3,2,18)$      & -\\
      Conv1D          & $(1 \times 3,2,18)$      & Dropout $20\%$ $ReLU$ \\
      Conv1D          & $(1 \times 3,2,50)$      &-\\
      Global Avg. Pool& $(1 \times 2,2,1)$       & - \\
      Feed-forward    & $(-,-,15) $               & $Leaky\,ReLU$ \\
      \hline
      \hline
    \end{tabular}
\end{table}

The edge update function is a feed-forward neural network. The input of the edge function is 
the temporal difference between observations. Both networks cast their inputs to vectors of the same size.
The $GN_{core} :\{ \phi^{u_c}, \phi^{e_c}, \rho^{e \rightarrow u} \}$ network, consists of small feed-forward neural networks 
for the node MLP $\phi^{u_c}$ and the edge MLP $\phi^{e_c}$. The input of the edge MLP is the sender and receiver state and 
the previous edge state. The MLP is implemented with a residual connection to allow for better propagation of gradients 
through multiple steps \cite{resnet}. 
\begin{equation*}
  \mathbf{e}'_i \leftarrow \phi^{e_c}(\mathbf{e}_i, \mathbf{u}_{s_i}, \mathbf{u}_{r_i}) = \bar\phi^{e_c}(\mathbf{e}_i, \mathbf{u}_{s_i}, \mathbf{u}_{r_i}) + \mathbf{e}_i
\end{equation*}
In this work, the $Mean(\cdot)$ aggregation function was chosen, which does not depend strongly on the in-degree of the state nodes $Z_i$ (i.e. number of incoming messages) which corresponds to step $2$ in algorithm \autoref{alg:gnnblock}.
\ifthenelse{\boolean{extended}}%
{The $Sum(\cdot)$ aggregation function is expected to be
influenced strongly by the number of incoming number of edges to a node, and since we aim 
for uniform treatment of past edges it was not considered. 
  \footnote{The other possible simple aggregation functions,$Max(\cdot)$ and  $\Min(\cdot)$, were tested and performed worse in preliminary numerical experiments}}%
{}%
The node MLP of the core network is also implemented as a residual MLP.
\begin{equation*}
  \mathbf{u'}_i\leftarrow \phi^{u_c}(\mathbf{u}_i, \mathbf{\bar{e}}_i) = \bar\phi^{u_c}(\mathbf{u}_i, \mathbf{\bar{e}}_i) + \mathbf{u}_i
\end{equation*}

The $GN_{core}$ network is applied multiple times to the output of $GN_{enc}$. This ammounts to the shared weights variant of GNs which 
allow for propagation of information from multiple steps while costing a small memory footprint.
After the last $GN_{core}$ step is applied, a final graph-independent layer is applied. At this point, for further computation 
only the final state of the last node is needed, which is the one corresponding to the last observation. The state of the last node 
is passed through two MLPs that terminate with $softplus$ activation functions
\begin{equation}
    Softplus(x) = log(exp(x) + 1).
\end{equation}
The $Softplus$ activation is needed for forcing the outputs to be in $(0,+\inf)$, since they are used as parameters for a $Gamma$ distribution which in turn is used to
represent the RUL estimates. The GraphNet computation procedure detailed above is denoted as 
\begin{equation}
  g_{out} = GN_{tot}(g) = GN_{dec} \circ GN_{core}^{(N_{c})} \circ GN_{enc} (g_{in})
\end{equation}
where $GN_{core}^{(N_{c})}$ denotes $N_{c} $ compositions of the $GN_{core}$ GraphNet and ``$g_{in},\,g_{out}$'' are the input and output graphs. 
The vertex attribute of the final node is in turn used as rate ($\alpha(GN_{tot}(g_{in}))$) and concentration ($\beta(GN_{tot}(g_{in}))$) parameters of a $Gamma(\alpha, \beta)$ distribution.
For ease of notation, the parameters (weights) of all the functions involved are denoted by ``$\bm{\theta}$'' and the functions that return the rate and concentration 
are denoted as $f_{\alpha;{\mathbf{\theta}}}$ and $f_{\beta;{\mathbf{\theta}}}$ correspondingly to denote explicitly their dependence on ``$\bm{\theta}$''. 
The $Gamma$ distribution was chosen for the distribution of the output values since they correspond to remaining time and they are necessarily positive. 
The GN described above is trained so as to maximize directly the expected likelihood of the remaining 
useful life estimates. For numerical reasons, equivalently, the negative log-likelihood (\textbf{nll}) is maximized. 
The optimization problem reads,

\newcommand{\Expect}{\mathbb{E}_{(\mathcal P, \mathcal{S})}}
\begin{align*}
  &\argmax_{\bm{\theta}} \Expect[p(\mathbf{y} | \mathbf{g})]  \propto \argmax_{\bm{\theta}}  \prod_{i=1}^{N^{s,p}} p(y_i | g_i)  \equiv \\
  &= \argmin_{\bm{\theta}} \sum_{i=1}^{N^{s,p}}\Big(-\log p(y_i | f_{\alpha;\bm{\theta}}(g_i),f_{\beta;{\bm{\theta}}}(g_i) )\Big) \numberthis 
  \label{eqn:nll}
\end{align*}
where $\mathbf{g}$ corresponds to the sets of input graphs, and $\mathbf{y}$
corresponds to the estimate of RUL for the last observation of each graph.
The input graphs in our case consist of nodes, which correspond to
observations and edges with time-difference as their features.
Correspondingly $g_i$ and $y_i$ are single samples from the aformentioned set of causal 
graphs and remaining useful life estimates and $N^{s,p}$ denotes the number of 
sampled causal graphs from experiment $p$ used for computing the loss (i.e. \textit{batch size}).
The expectation symbol is approximated by an expectation over the 
set of available training experiments denoted as $\mathcal P$ and the 
random causal graphs created for training
$\mathcal S $.
The gradients of \autoref{eqn:nll} are computable through \textit{implicit re-parametrization gradients} \cite{figurnov2018implicit}. 
This technique allows for low-variance estimates for the gradient of the nll 
loss with respect to the parameters of the distribution, which in turn allows for a complete end-to-end differentiable 
training procedure for the proposed architecture. 

As in recurrent neural network models \cite{gru,lstm}, and \cite{wavenet}, a gated-tanh activation function 
was used for the edge update and node update core networks.
\begin{equation*}
	h(y)=sigmoid(W_g y) \odot tanh(W_a y)
\end{equation*}
GraphNets using this activation strongly outperformed the ones using tanh but showed 
similar performance to the ones using relu activation. 
Networks for the edge and node MLPs were tested with widths $30,\,50,$ and $100$. The smaller networks tested (size $30$) consistently outperformed networks with size $50$ and for the most part had similar performance with networks with size $100$ for some cases. The $30-$unit networks were selected for the presented results.

\subsection{Recurrent Neural Network with Temporal CNN Feature Extractors (LSTM-tCNN)}
\begin{figure}[h!]
  \center
  \includegraphics[width=0.35\textwidth]{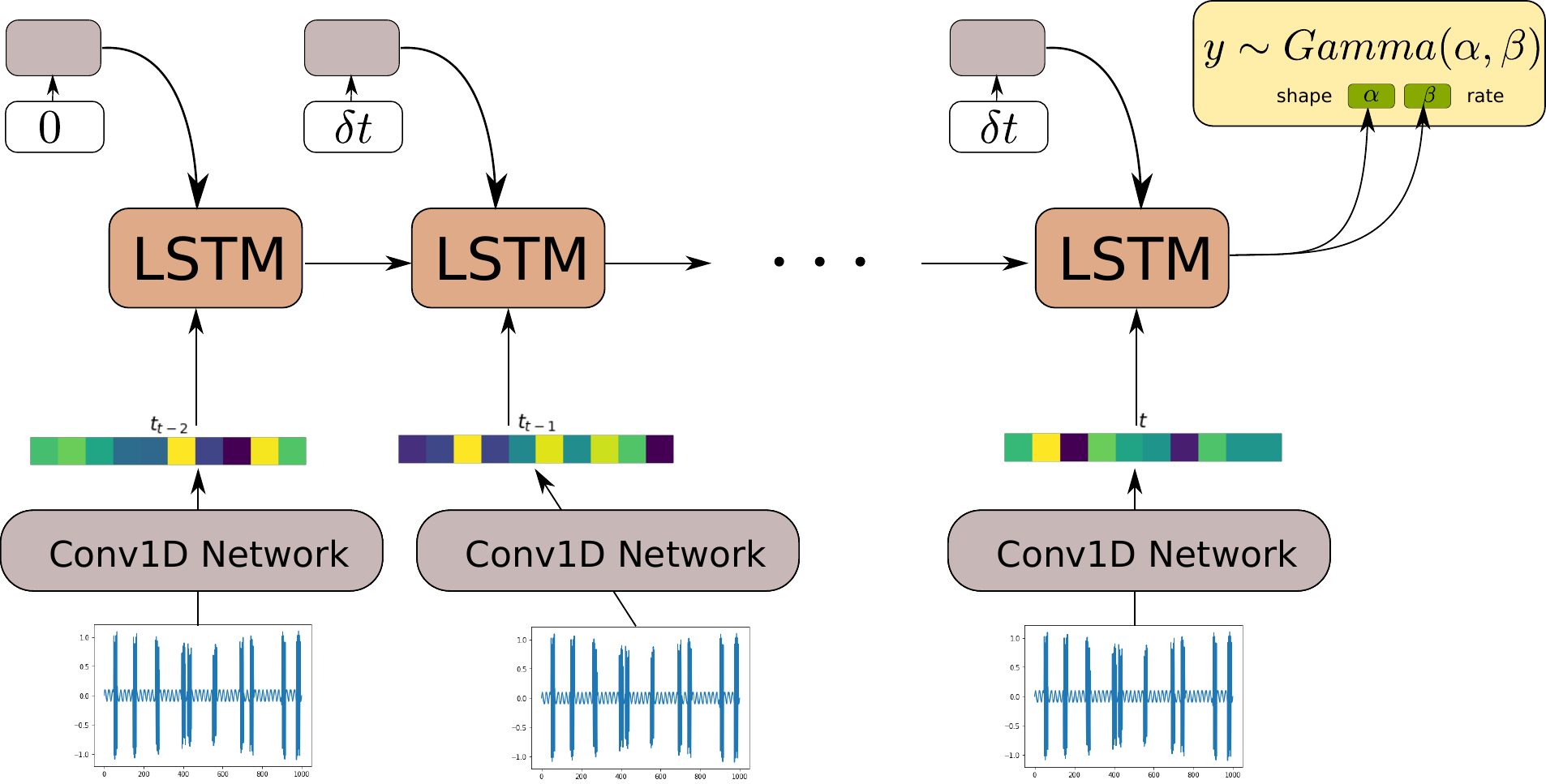}
  \caption{Model architecture of LSTM-tCNN. }
  \label{fig:ltrtcnn}
\end{figure}

The Causal GNN component of the architecture detailed in \autoref{sec:gnncomputation} is used to satisfy the following desiderata:
\textbf{(1)} to allow for computationally efficient and parallelized propagation of information from time-instants in the 
distant past with respect to the current timestep and \textbf{(2)} to allow for learning a state-propagation 
function and hence dealing with arbitrarily spaced points in a consistent manner. Although gated RNNs, such as GRUs and LSTMs,
rely on sequential computation between timesteps, and therefore less parallelizeable, they are known to be relatively effective in 
dealing with long dependencies. Moreover, by appending the time difference between observations in the input gate of the RNN
the RNN can learn to condition the predictions for the propagated state not only on the previous state and the CNN feature extractor 
input but also to the time-difference between different RNN steps \cite{zhu2017next}. One such model, using an LSTM cell, is depicted in \autoref{fig:ltrtcnn}.



\ifthenelse{\boolean{extended}}
{
\subsection{Sampling for Training \& Evaluation}
As mentioned above, the computational efficiency of the proposed technique 
relies to a certain extent on sampling from past observations, rather than 
retaining all of them and processing them in a chain-like manner. 
A custom sampling procedure was implemented, where a small number of past observations 
is used. The observations are sampled to belong on a time-window of 2000s. Since the network 
is to be used for arbitrary number of observations, a schedule of $1, 2, 5$ and $10$ past 
observations was used during training, with the number of observations rotated for each epoch. 

The computational cost of each message-passing step of the GNN-tCNN evaluation, increases as $n^2/2$ where $n$ is the number of points incorporated in the prediction (when considering all 
the edges between all the past observations). Since the computational cost increases 
quadratically with the number of points, a relatively small number of past points was used.
A simple technique to alleviate this problem is, for instance, 
to sample at random without replacement or with replacement (i.e. \textit{bootstrapping}) 
from the past observations is one such technique. 

The sampling scheme used in the presented results is such that the observation points are chosen 
so as not to be too close. Namely, points are at least 100s apart. During training, 20\% of the sampled graphs are used as a validation step. The network is trained with linear burn-in for 10 epochs, the 
Adam optimizer (learning rate $0.001$) and a learning rate decay rate of $0.99$ after the 40th epoch for a maximum of 300 epochs.
Early stopping was performed when the validation set error did not decrease for 50 epochs.
}{
}
\section{Results}
\subsection{Simulated Dataset}
Although it is easy to create a large number of training and test set experiments from the simulated dataset, in 
order to keep the simulated use-case realistic, only 12 experiments were used for training and a set of 3 experiments 
were used as a test set. Representative prediction results for the test-set experiments are shown in \autoref{fig:resultsfict}.
The accuracy of the model is inspected in terms of the expected negative log likelihood (smaller is better).
When more observations are used, the estimates for the RUL of the fictitious processes are more accurate for a larger portion of the entirety of the 
observations. When a single observation is used, which completely neglects the long-term evolution of damage but uses short-term features 
extracted by the learned graph-independent $\phi^{u_0}$ which is a temporal CNN, the RUL estimates are inaccurate and fluctuate 
at the beginning of the experiment (top-right side of the first column of the plots). It is also observed that when using more observations, although 
the uncertainty bounds are not becoming smaller at the beginning of the fictitious experiment, the estimates of the trend of degradation, and consequently the 
remaining useful life of the component are accurate and smooth early on in the course of the experiment.

This observation provides evidence that the proposed architecture accurately captures both features of the high-frequency time series 
through the CNNs of the first graph-independent processing step, and the long-term evolution of the time series through the GraphNet processing steps.
Moreover, the estimated probability distributions of RUL become more concentrated closer to failure, while they are wider at the beginning of the 
experiments. This observation aligns with the intuition that it is not possible to have sharp estimates in the beginning of the experiments.
%
%
\subsection{Bearings Dataset}
\begin{figure}
    \centering
  \subfloat[\textbf{GNN-tCNN}, Simulated\label{fig:resultsfict}]{%
    \includegraphics[width=0.45\textwidth]{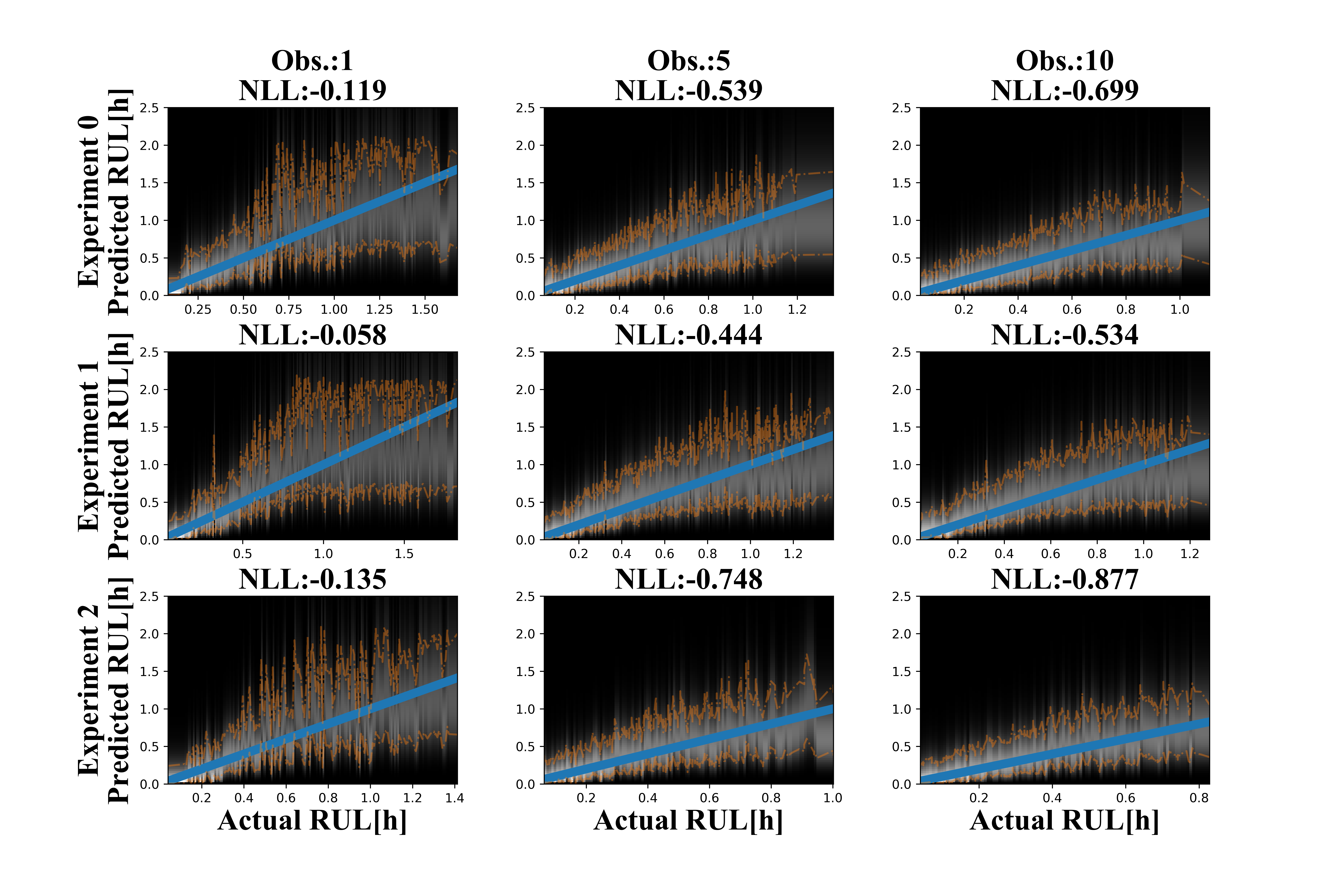}}
    \\
  \subfloat[\textbf{GNN-tCNN}, Bearings\label{fig:bearingsresults}]{%
    \includegraphics[width=0.45\textwidth]{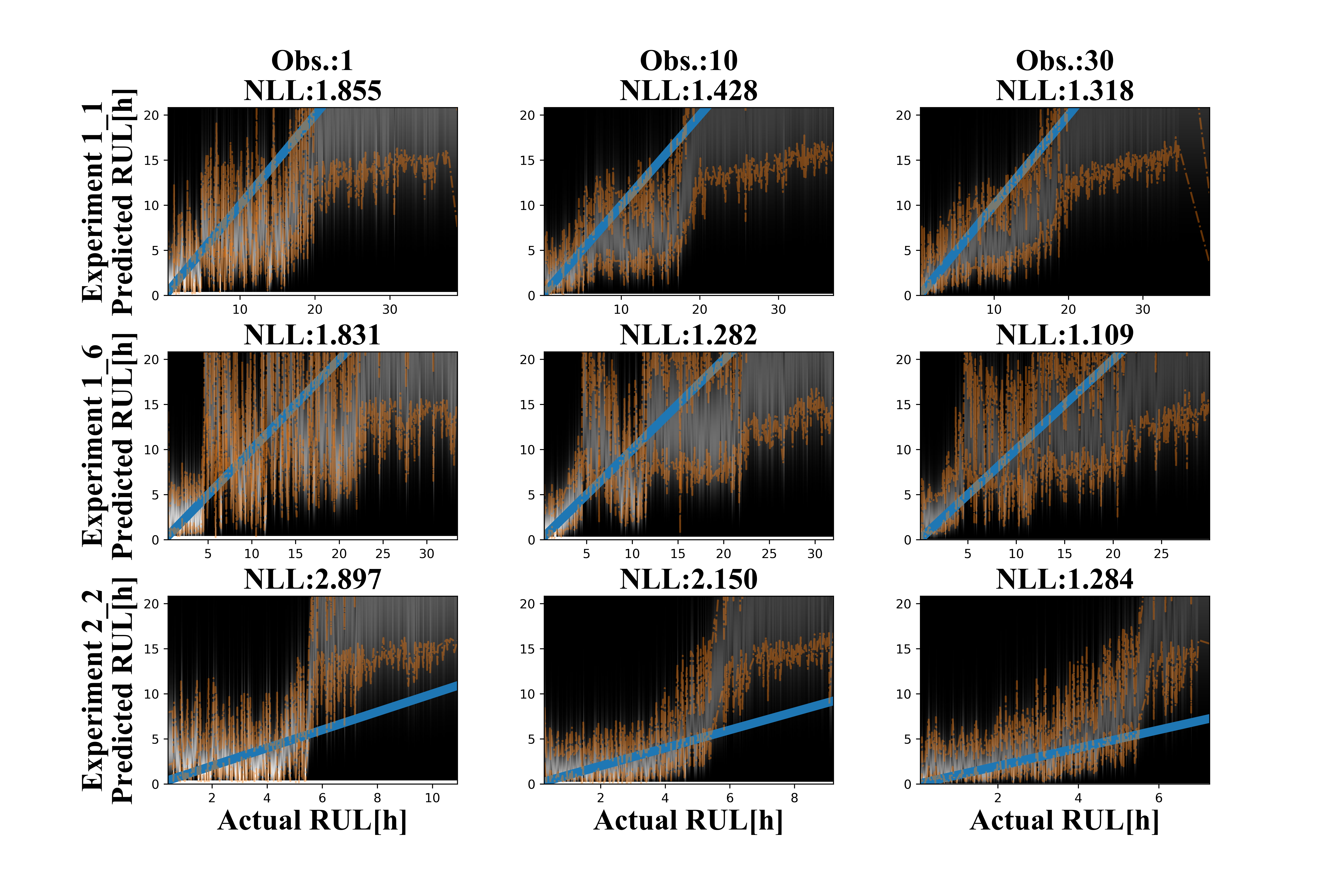}}  
    \\
  \subfloat[\textbf{LSTM-tCNN}, Bearings\label{fig:ltrcnn}]{%
    \includegraphics[width=0.45\textwidth]{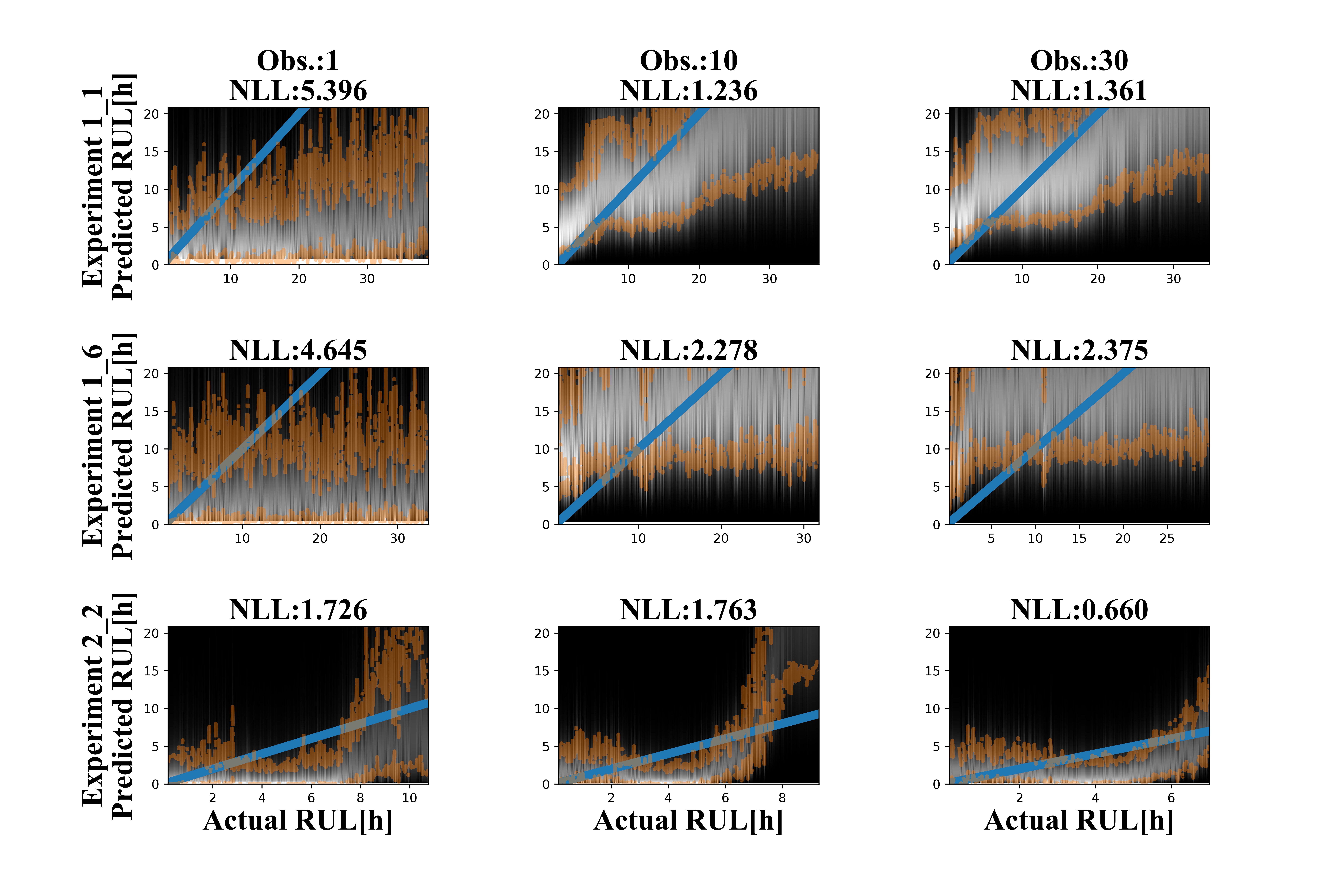}}
  \caption{(a) Results on the simulated dataset using GNN-tCNN. (b), (c) Results on the bearing dataset using GNN-tCNN and LSTM-tCNN.}
  \label{fig:figstogether}
\end{figure}

Results for representative experiments from the test-set of the FEMTO-bearing dataset are shown in \autoref{fig:bearingsresults}. Similarly to the simulated experiments, predictions 
are characterized by smaller uncertainty closer to failure and the degradation trends are captured effectively.
Predictions employing up to 30 arbitrarily spaced observations from the past 2000 seconds are shown. Using more observations (more than 30) did not signifficantly improve the 
accuracy or the uncertainty bounds of the predictions. This may be due to the fact that the damage phenomenon is slowly evolving, thus using a larger number of points does
not offer more information on the evolution of the phenomenon. 

%


\section{Conclusion}

Two neural network architectures (GNN-tCNN and RNN-tCNN) were applied to the problem of remaining life 
assessment on real-world experimental data and on artificial data from a stochastic 
degradation process. Both architectures seem to capture the long-term dependences
on features of the considered time series. The GraphNet architecture proposed features a 
causal connectivity structure that can capture with less sequential computation long-term
dependencies in the time series. Finally, an effective gradient-based technique, which employs 
low-variance reparametrization-based gradient estimators for fitting distributions with positive 
quantities of interest, such as RUL estimates, was employed. The proposed architectures 
are intuitive and easy to implement.\footnote{Code to reproduce the experiments in the paper will be made available upon publication}

Although only RUL estimation problems were considered in this work, 
the non-sequential causal approach to dealing with long-term dependencies may be applicable to 
further applications where non-regularly sampled time series arise (e.g. electronic health records).
In future works, other time-series tasks such as time-series generation or unsupervised/self-supervised 
learning \cite{hyvarinen2016unsupervised} are to be attempted, employing the 
GNN-tCNN architecture.

\bibliographystyle{Bibliography/IEEEtranTIE}
\bibliography{Bibliography/bibliography}\ 
\vspace{-1cm}
\begin{IEEEbiography}[{\includegraphics[width=1in,height=1.25in,clip,keepaspectratio]{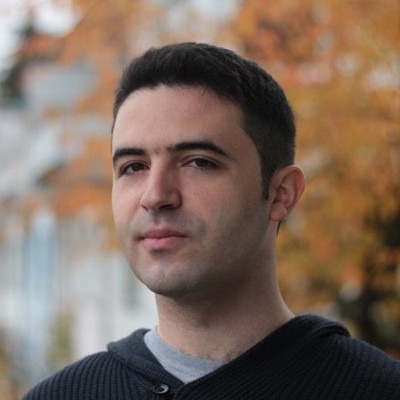}}]
{Charilaos Mylonas} was born in Thessaloniki, Greece. He holds a Dipl. Ing. Structural Engineering degree from Aristotle University of Thessaloniki and a MSc on Computational Science and Engineering from ETH Z\"urich.

  He previously worked as a scientific software developer at ETH Z\"urich and as a full-stack software developer in the banking sector. He is currently working towards his Ph.D. on the topic of remaining life assessment and uncerainty quantification for wind turbines. 
\end{IEEEbiography}
\vspace{-1cm}
\begin{IEEEbiography}[{\includegraphics[width=1in,height=1.25in,clip,keepaspectratio]{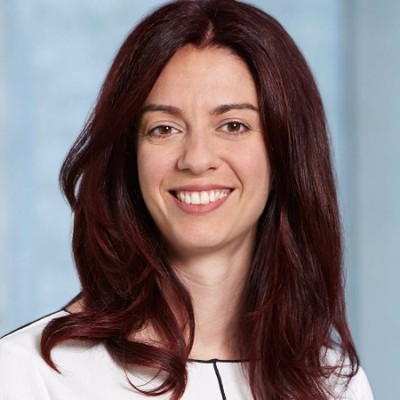}}]
{Eleni Chatzi}  was born in Athens, Greece.  She received her PhD (2010) from the Department of Civil Engineering and Engineering Mechanics at Columbia University. She is Chair of Structural Mechanics and Monitoring at the Department of Civil, Environmental and Geomatic Engineering of ETH Zürich. Her research interests include the fields of Structural Health Monitoring (SHM) and structural dynamics, nonlinear system identification, and intelligent assessment for engineered systems. She is leading the ERC Starting Grant WINDMIL on smart monitoring of Wind Turbines. Her work in the domain of self-aware infrastructure was recognized with the 2020 Walter L. Huber Research prize, awarded by the American Society of Civil Engineers (ASCE).


\end{IEEEbiography}

\end{document}